\documentclass{IOS-Book-Article}

\usepackage{mathptmx}
\usepackage{soul}\setuldepth{article}
\usepackage{graphicx}
\usepackage{amsmath}

\def\hb{\hbox to 11.5 cm{}}

\newcommand{\ST}[1]{}
\newcommand{\AB}[1]{}
\newcommand{\AP}[1]{}
\newcommand{\FG}[1]{}
\newcommand{\LE}[1]{}
\newcommand{\MR}[1]{}

\begin{document}

\pagestyle{headings}
\def\thepage{}

\begin{frontmatter}              

\title{Lifelong Personal Context Recognition%
\thanks{The work by Xiaoyue and Haonan is funded by the China Scholarships Council. The work by Andrea Bontempelli, Fausto and Marcelo was partially funded by the project “DELPhi - DiscovEring Life Patterns” funded by the MIUR (PRIN) 2017. The research of Stefano and Andrea Passerini was partially supported by TAILOR, a project funded by EU Horizon 2020 research and innovation programme under GA No 952215.}}

\markboth{}{June 2021\hb}

\author[]{Andrea Bontempelli, Marcelo Rodas Britez, Xiaoyue Li, Haonan Zhao}
\author[]{Luca Erculiani, Stefano Teso, Andrea Passerini, Fausto Giunchiglia}
%
%
%
\runningauthor{Bontempelli et al.}
\address{University of Trento, Italy}

\begin{abstract}
    We focus on the development of AIs which live in lifelong symbiosis with a human. The key prerequisite for this task is that the AI understands - at any moment in time - the \textit{personal situational context} that the human is in.
    We outline the key challenges that this task brings forth, namely \textit{(i)} handling the human-like and ego-centric nature of the the user's context, necessary for understanding and providing useful suggestions, \textit{(ii)} performing lifelong context recognition using machine learning in a way that is robust to change, and \textit{(iii)} maintaining alignment between the AI's and human's representations of the world through continual bidirectional interaction.
    In this short paper, we summarize our recent attempts at tackling these challenges, discuss the lessons learned, and highlight directions of future research.  The main take-away message is that pursuing this project requires research which lies at the intersection of knowledge representation and machine learning. Neither technology can achieve this goal without the other.
\end{abstract}

\end{frontmatter}
\markboth{June 2021\hb}{June 2021\hb}

\vspace{-0.2cm}
\section{Introduction}

We focus on the development of AIs which live in lifelong symbiosis with a human and interact with her via smart wearables, for instance, smart phones, smart watches, or also - for certain health-centered applications - medical devices. 
By \textit{Human-AI Symbiosis} we mean here an AI which is aware of the user's life and well-being, in \textit{all aspects of the user's everyday life}, this being the premise for the AI to provide added value services. It is therefore a \textit{holistic} symbiosis enabling a level of interaction which goes far beyond what achieved so far by personal assistants, which focus on single aspects of the user's life, such as calendar and agenda management, physical activities and fitness, education and learning, information access and management, cf.~\cite{mitchell1994experience,guha2015user,dempsey2017teardown}. This is a very difficult task whose underlying difficulty, still unsolved, has been known for decades. As discussed in John McCarthy's Turing Award lecture \cite{mccarthy1987generality}, as soon as one gets out of specialized and well defined domains, AIs suffer from the problem of \textit{brittleness}, namely their inability to deal with situations differing even marginally from the scope for which there they were devised. McCarthy called this the \textit{Problem of Generality}. The original formulation was meant only for knowledge-based approaches. For some time it looked like machine learning could provide an effective solution to this problem. However, as discussed in some detail in \cite{gini2018artificial}, this turned out not to be the case. As of today, machine learning, and deep learning in particular, are hardly usable in applications which are not restricted in scope or not time-invariant.

The key intuition underlying our approach is that, in a world which is ever changing and which does so in a largely unpredictable way, the only way for an AI to avoid the problem of brittleness is to make sure that its understanding of what is happening is completely aligned with that of its user.
In fact, in a world where \textit{(i)} in time, nothing is ever equal to itself and where \textit{(ii)} the meaning of what is happening, i.e., the \textit{intended semantics} of any representation of what is happening, is in the mind of humans and, as a particular case, in the mind of its reference user, humans are an AI's ultimate source of information.
In turn, the \textit{only} possible way to carry out this objective is to build an AI that, as a lifelong effort, is able to continually interact with and get relevant information from its reference user, in particular when something new and unforeseen happens.
This type of alignment requires full mutual and \textit{bidirectional} explainability between the AI and her reference user. The direction from the user to the machine is what is usually done in Knowledge Representation (KR) and Machine Learning (ML). The other direction has recently emerged under the moniker of Explainable AI (XAI)~\cite{Gui2018,Miller2018} as a major requirement to achieve human-centric and trustworthy AI. The bidirectional aspect of explainability, so far largely unstudied, is however equally and maybe even more important.
Via this type of interaction, the AI will become able of acquiring that type of flexibility which allows humans to understand other humans, and also what happens in the real world, even in presence of differing or wrong information.

Based on the above assumptions, the research described here is based on three main ingredients:
\textit{(i)} A general KR mechanism for defining the personal context so as to enable the machine to view the world in user's terms.
\textit{(ii)} A suite of general ML mechanisms enabling an AI to perform context recognition in the wild, adapting to changes in the world and in its user.
\textit{(iii)} A lifelong \textit{machine-human alignment loop} to maintain alignment through bidirectional interaction.
The rest of the paper will describe the progress made in the three lines of research mentioned above (Sections \ref{s1}, \ref{s2}, \ref{s3}). Section \ref{s4} will report the lessons learned also indicating the implications on the way ahead.
Most of this work has been applied in real world scenarios, based on data collected during experiments designed as part of this research. The SmartUnitn2 dataset, built with an experiment involving 158 university students over a period of four weeks (see, e.g., the description in~\cite{KD-2017-PERCOM}) will be the base for the examples and the results described in the following.

\vspace{-0.2cm}
\section{Representing the personal situational context}
\label{s1}

\begin{figure}[!htb]
\vspace{-0.3cm}
\includegraphics[width=0.9\linewidth]{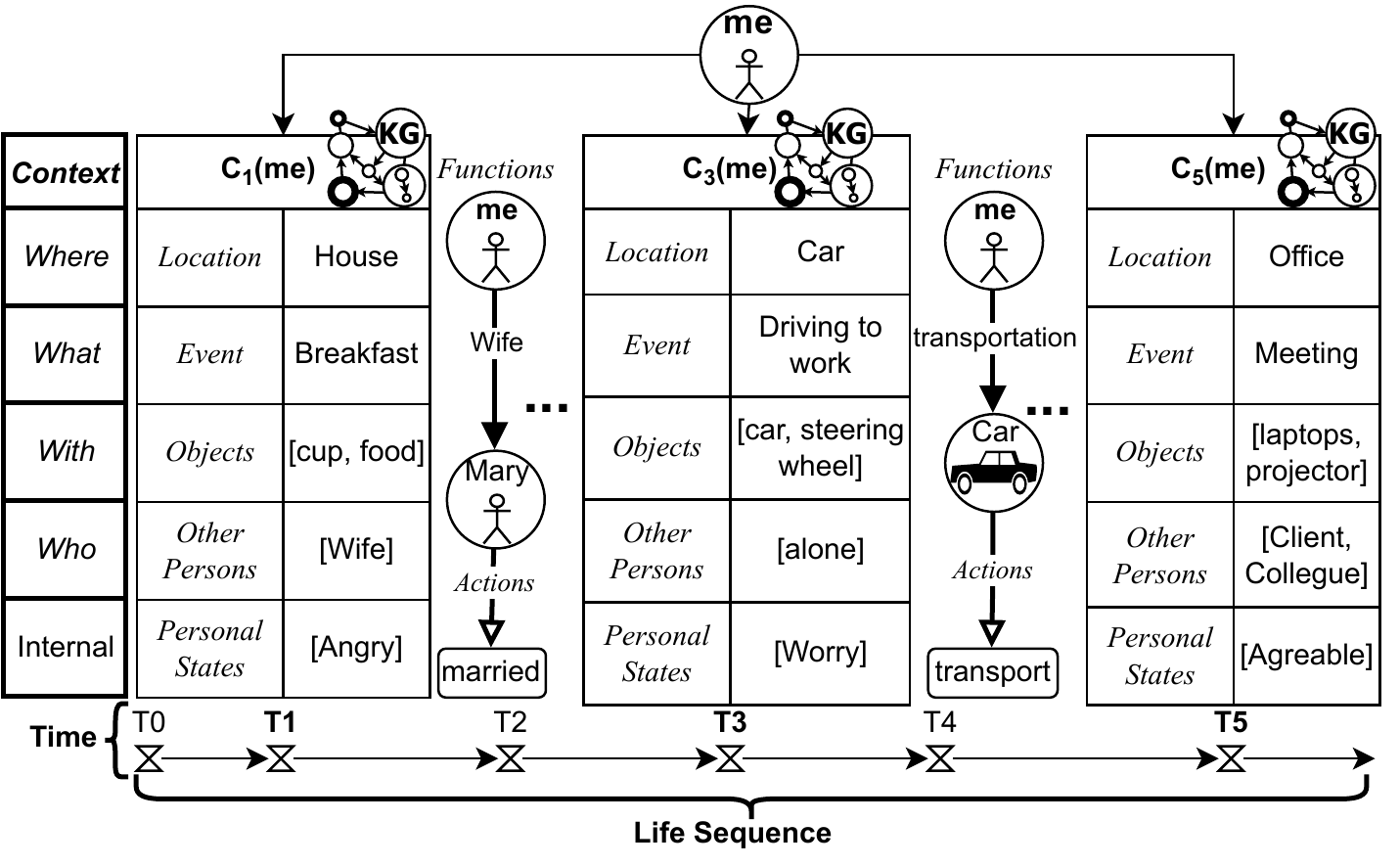}
\vspace{-0.2cm}
\caption{A three context life sequence of \texttt{me}.} 
\vspace{-0.3cm}
\label{fig:context}
\end{figure}

The notion of context used here was originally (informally) defined in~\cite{KD-1993-giunchiglia} as \textit{``a theory of the world which encodes an individual’s subjective perspective about it''}. The key intuition underlying the use of contexts is that generality is achieved by moving from the approach where there is only one monolithic theory of the \textit{objective} world to an unbound number of \textit{subjective} views, modeled as contexts, each providing a partial view of the world, i.e., the set of facts which are locally relevant to the current activity~\cite{KD-1995-bouquet}.
In this work, we represent contexts as Knowledge Graphs (KGs) and assume that the AI is able to store any number of them as (sets of) \textit{Life Sequences}~\cite{KD-2017-PERCOM}, where a Life Sequence is a sequence of contexts. Figure~\ref{fig:context} represents a small life sequence with three contexts.

\noindent
Informally, the \textit{personal situational context} describes the circumstances of a person in terms of space-time (Locations and Events), internal context (Personal States), social context (Other Persons), object environment (Objects), and functional relations (Functions and Actions) between the components. Here we talk of \textit{situational} context to mean the context within which a person operates in any moment of time.
The situation context is composed by the \textit{Location} of \texttt{me}, which defines the spatial boundaries, and the \textit{Event} within which \texttt{me} is involved at the moment. An event is parameterized on the location as we may have different events occurring in the same location. Location and event are the priors of experience, defining the scenario that needs to be modeled. The change of context coincides with a change in the current location or in the current event.

Given the notion of context as from above, we define the notion of \textit{Life sequence} as a sequence of contexts during a certain period.
We assume that \texttt{me} is involved in only one personal context at a time. In fact, at any given moment, a person can be in only one place.
This context representation has been used in \cite{giunchiglia2021streaming,KD-2022-Xiaoyue}.

\vspace{-0.2cm}
\section{Continually evolving context recognition}
\label{s2}

Given the notion of personal context, the question becomes how to obtain context information in applications where it is needed.
The personal context of a user is typically not directly accessible, at least not in real-time, so it must be inferred from what other information is available.
What is needed is a mechanism for enabling an AI to carry out this step.  This problem is what we refer to as \textit{personal context recognition} (PCR).

\paragraph{Context recognition in a static world.}  From one perspective, PCR can be viewed as a generalization of tasks like activity recognition~\cite{chen2021deep} -- where one is given access to measurements from handheld or wearable sensors and has to derive what (unobserved) action the user is performing -- from a single aspect of the personal context (namely, activity) to the whole context.
This immediately suggests an ML approach consisting of two steps:  collecting examples of sensor recordings annotated with context information and then using this data to learn a map between the two that generalizes to unseen situations.
A recent investigation of learning techniques for PCR carried out on the SmartUnitn2 data shows that indeed automated recognition can be achieved with some degree of success~\cite{shen2020multi,zhang2021putting}.

Figuring out what predictors and architectures are best suited for this task will likely involve borrowing ideas from activity recognition and related areas, while mixing in strategies for dealing with specific aspects of PCR, such as incrementality and support for interaction, which are critical for lifelong alignment (as discussed below).
Another important element is that the personal context is inherently \textit{structured}~\cite{zhang2021putting}, in the sense that its various aspects are correlated -- e.g., a person's activity is strongly influenced by the location that she is in -- and constrained by the structure of the context knowledge graph.  This hints at the need for developing or repurposing structure-aware predictors.
Few or no architectures satisfy all these desiderata and achieve high performance, although progress in this direction is being made~\cite{teso2022efficient}.

So far, PCR can be viewed as a rather standard (although highly non-trivial) ML task under KR constraints.
The real challenges appear when moving to the lifelong setting.
Delivering high recognition performance over time requires AIs to be robust to changes \textit{in the world} and \textit{in their user}, which are arbitrary and essentially impossible to anticipate~\cite{dietterich2017steps}, and therefore to be:
\textit{(i)} incremental, so as to promptly update as new information is acquired,
\textit{(ii)} able to autonomously detect changes and adapt by acquiring appropriate feedback, and
\textit{(iii)} able to exploit the prior information stored in the previous contexts, while at the same time, being able to fill them with the knowledge just learned.
We unpack these elements next.

\paragraph{The user's description of the context changes.}  Over time, the user describes the same world in different ways, even if it is not changed.  For instance, the user may refer to the same park as central park, urban park, or municipal gardens.  At the same time, a relevant fraction of the annotations supplied by users to describe his/her context are unreliable due to mistakes or inattention~\cite{zeni2019fixing}. Given that acquiring new annotations is expensive and their amount is limited, this issue badly affects the performance of PCR predictors.
\textit{Skeptical learning}~\cite{zeni2019fixing} is a recent interactive learning strategy that tackles annotation inconsistencies. The machine asks the user to revise her annotations if it is confident that there is an inconsistency~\cite{zeni2019fixing,zhang2022skeptical}. The machine's uncertainty can be estimated using Bayesian~\cite{bontempelli2021learning} or frequentist~\cite{teso2022efficient} techniques, while enabling the AI to explain its skepticism by showing past examples that support the model's  suspicion~\cite{teso2021interactive}. Both past and new data can be fixed by the user leading to better data and models~\cite{teso2021interactive}.
This helps the AI to ensure that the information it stores is globally consistent.\footnote{The requirement of consistency was already identified in \cite{mccarthy1987generality}. McCarthy proposed \textit{non-monotonic reasoning} as a general consistency-preserving mechanism in a changing world. Later, \cite{KD-1988-giunchiglia} proposed an approach to non-monotonicity based on the use of contexts. However, these and more recent works assume that the user would provide the new information, hindering scalability in practical applications.}

\paragraph{The world itself changes.}  Over time the world changes, and so does the user and her understanding of it. E.g., if the user has to relocate, the structure of her personal context changes.
From a statistical perspective, this can be viewed as a form of \textit{concept drift}~\cite{tsymbal2004problem,gama2014survey} and as such it affects the distribution of sensor observations and of personal context given observations.  For instance, during the semester, our user may spend most of her time studying at the library, but once the finals are over, she stops going to the library as often and while there she is less likely to be studying.
Here, however, drift is more complex as it can affect the knowledge encoded in the context knowledge graph.  In other words, whereas concepts like ``Friend'' and ``Library'' are essentially immutable, the \textit{specific} friends and libraries that matter to the user do change over time, for instance when she graduates.  We refer to this as \textit{knowledge drift}~\cite{bontempelli2021human}:  concepts and relations can become obsolete or irrelevant, and new ones may need to be acquired.
Failure to align the AI's understanding to the updated knowledge may lead to providing useless or actively harmful predictions.
Given the ego-centric nature of the personal context, the only way to counter knowledge drift is to interact with the user herself.  This is actually necessary:  different forms of knowledge drift leave a similar footprint on the data, which is thus insufficient to disambiguate between -- and therefore properly adapt to -- them.
In \cite{bontempelli2021human}, we developed a novel algorithm that tackles all of these issues by integrating \textit{automated} drift detection and adaptation of the machine's knowledge graph with an \textit{interactive} step in which the user helps the machine to disambiguate between alternative kinds of knowledge drift.

\vspace{-0.2cm}
\section{Machine-human alignment loop}
\label{s3}

The goal here is to design a general \textit{machine-human alignment loop} that ensures alignment over the life span of the AI.
We posit that doing so will involve \textit{continual}, \textit{bi-directional interaction}.
The question is then how to structure it such that it is cognitively cheap (on the human's side) and computationally affordable (on the machine's side).
This is where most future research lies. For instance, an important issue is how to make sure that the user does not get bothered by a (life)long intensive interaction with the AI (most often not generated by her). A second main issue is that the ML techniques above generate a high number of heterogeneous questions, asking for very different information, motivated by different purposes and based on different background knowledge. Even assuming that the user does not get bothered, how can we make sure that the user does not get confused therefore providing the AI with wrong information? 

A second order of problems is that, in order to avoid brittleness, we assume the latter to be based on the context representation in Section~\ref{s1}, an assumption which requires the ability of translating all the possible outcomes of ML in KR. Solving this problem presents a number of difficulties:
\textit{(i)} The meaning of different ML labels may be related (e.g., they can be synonyms, or have more/less general meanings, or even be from different natural languages).
\textit{(ii)} Labels are polysemous and, as such, have multiple meanings that need to be disambiguated.
\textit{(iii)} Labels may be unknown to the AI which therefore needs to extend its vocabulary.
\textit{(iv)} The assumption of using labels rather than full text is quite limiting.
Some simple versions of this problem have been dealt by the Skeptical Learning~\cite{zhang2022skeptical} and are based on the use of a large multilingual resource, called UKC~\cite{KD-2017-IJCAI}, but we are just at the beginning. The above is at the level of language. But problems exist also at the knowledge level in that a label, even the same label in a different context, may stand for (in the context knowledge graph) for an entity, a class name or property name or value. There is finally a last complexity in that, independent of its meaning, the label learned in a certain moment holds only for that moment, while the local context must represent meaning, and how it is partially stable in time, independently of the local fluctuations. Some very initial work is described in~\cite{giunchiglia2021streaming}.

On the machine side, the need to obtain supervision can be addressed by adapting machine-initiated and human-initiated interaction protocols, such as active learning~\cite{settles2012active} guided learning~\cite{attenberg2010label}.  In real-world lifelong alignment interaction however entails solving additional problems, such as choosing \textit{when} to request annotations so as to maximize response time and quality.
Moreover, in order to handle change, the machine will have to interleave requests for supervision with skeptical learning \textit{and} knowledge drift detection and adaptation.
How to properly model and implement bidirectional interaction of this kind is left to future work.

\vspace{-0.2cm}
\section{Lessons learned}
\label{s4}

What described above are only first steps. Still a few general lessons have already been learned which provide guidelines for the work to come.

The first relates to the KR dimension of the problem. Implementing contexts presents two non trivial difficulties. The first is that the notion of context should allow for the representation of \textit{any} possible real world situation, thus effectively achieving the requirement of generality. The second is that of representing and preserving identity across different contexts. For instance, it must be possible to represent the set of entities, e.g., people, locations, events, which occur across multiple contexts, during the lifetime of a person. Furthermore this must be done in a way that, in such contexts these entities are described by different properties and different, possibly contradicting, property values, this being the key for modeling non monotonic reasoning in a changing world~\cite{KD-1988-giunchiglia}.

The second relates to the ML dimension of the problem. Here there is obviously a lot to do in the improvement of the state of the art. But the two issues which seem more pressing and leading to new avenues relate to the full embedding of ML in time (which for instance makes the distinction between training and execution after training meaningless) and to the bi-directionality of the human-AI interaction. It is not a case that the first paragraph in Section \ref{s3} does not have citations. This is the exactly the empty space on which we are concentrating now.

The third relates to the Machine-Human alignment loop. The first wave of experiments provided clear evidence that there is a need to provide the AI with Computer Vision (CV) and Natural Language Processing (NLP) capabilities. Only CV allows the AI to construct a model of the world which is similar to that of the human and facilitates their mutual interaction.
NLP is needed to scale to the complexity of the world (see Section~\ref{s3}). Here the focus is on building interactions, e.g., in the form of simple dialogues, where the terms used by the AI have the meaning intended by the human.
Here the well known and still unsolved \textit{Semantic Gap} problem applies \cite{CV-2000-Smeulders-content}. The work described in \cite{erculiani2019continual,giunchiglia2021towards} is a first small step in this direction.

The fourth issue is the importance of running real world experiments in the wild. One reason is that there are no datasets about Human-Machine symbiosis. The second is that there is a need of ethics aware datasets, mainly because of its huge impact on the life of people. The third is that the evaluation of the AI we are developing, which evolves in time while, at the same time, changing the behaviour of the human, requires a careful design of the experiment. Various datasets have been collected, see \cite{SmartUnitnONE,SmartUnitnTWO,WeNetDiversityOne,WeNetDiversityOnePlus}, using the iLog platform~\cite{KD-2014-PERCOM,KD-2017-PERCOM}. A major challenge is the need of a general methodology to be used to run these experiments in a systematic way. This brings up the issue of inter-disciplinarity. Initially, the focus was on Philosophy and Cognitive Science, but here there is also a need to learn from the (Computational) Social Sciences (e.g., see \cite{KD-2020-Zeni1}).

\section*{Author profiles}
\begin{itemize}
    \item Andrea Bontempelli, PhD student, KR, ML and AI
    \item Marcelo Rodas Britez, Post-doctoral Researcher, KR and AI
    \item Xiaoyue Li, PhD student, KR and AI
    \item Haonan Zhao, PhD student, ML and AI
    \item Luca Eculiani, Post-doctoral Researcher, ML and AI
    \item Stefano Teso, Assistant Professor, ML and AI
    \item Andrea Passerini, Associate Professor, ML and AI
    \item Fausto Giunchiglia, Full Professor, KR, AI and ML
\end{itemize}

\bibliographystyle{vancouver}
\bibliography{references}
\end{document}